\documentclass[lettersize,journal]{IEEEtran}
\usepackage{amsmath,amsfonts}
\usepackage{algorithmic}
\usepackage{algorithm}
\usepackage{array}
\usepackage[font=normalsize,labelfont=sf,textfont=sf]{subfig}
\usepackage{textcomp}
\usepackage{stfloats}
\usepackage{url}
\usepackage{verbatim}
\usepackage{graphicx}
\usepackage{cite}
\usepackage{caption}
\usepackage{booktabs}
\usepackage{authblk}
\usepackage{flushend}
\begin{document}

\title{DexTac: Learning Contact-aware Visuotactile Policies via Hand-by-hand Teaching}

\author{
	Xingyu Zhang\textsuperscript{1,2},
	Chaofan Zhang\textsuperscript{1},
	Boyue Zhang\textsuperscript{1},
	Zhinan Peng\textsuperscript{3},
	Shaowei Cui\textsuperscript{1,\,*}\thanks{*Corresponding author: Shaowei Cui (shaowei.cui@ia.ac.cn)},
	and Shuo Wang\textsuperscript{1,2}
}

\affil{
	\textsuperscript{1} Institute of Automation, Chinese Academy of Sciences\\
	\textsuperscript{2} University of Chinese Academy of Sciences\\
	\textsuperscript{3}University of Electronic Science and Technology of China
}


\maketitle

\begin{abstract}
For contact-intensive tasks, the ability to generate policies that produce comprehensive tactile-aware motions is essential. However, existing data collection and skill learning systems for dexterous manipulation often suffer from low-dimensional tactile information. To address this limitation, we propose DexTac, a visuo-tactile manipulation learning framework based on kinesthetic teaching. DexTac captures multi-dimensional tactile data—including contact force distributions and spatial contact regions—directly from human demonstrations. By integrating these rich tactile modalities into a policy network, the resulting contact-aware agent enables a dexterous hand to autonomously select and maintain optimal contact regions during complex interactions. We evaluate our framework on a challenging unimanual injection task. Experimental results demonstrate that DexTac achieves a 91.67\% success rate. Notably, in high-precision scenarios involving small-scale syringes, our approach outperforms force-only baselines by 31.67\%. These results underscore that learning multi-dimensional tactile priors from human demonstrations is critical for achieving robust, human-like dexterous manipulation in contact-rich environments.
\end{abstract}


\section{INTRODUCTION}
In recent years, dexterous manipulation has captured increasing attention in various fields \cite{DexterityGen,DIH-Tele,DexCo,study,Progressive}. Robust dexterous manipulation depends on applying appropriate force and selecting an appropriate contact region (e.g., fingertip, finger pad, or finger side) during object interaction \cite{DexForce}. Suboptimal choices often lead to finger slippage and task failure. One effective approach to teaching robots how to apply appropriate forces within right contact regions is imitation learning (IL) \cite{ACT,mobileACT}, which leverages state-action pairs demonstrated by human experts to guide policies acquisition. However, how to collect high-quality visuotactile data and how to effectively learn contact-aware policies from human demonstrations for robots in contact-rich tasks remain challenging for dexterous manipulation.
\begin{figure}
	\centering
	\includegraphics[width=0.99\linewidth]{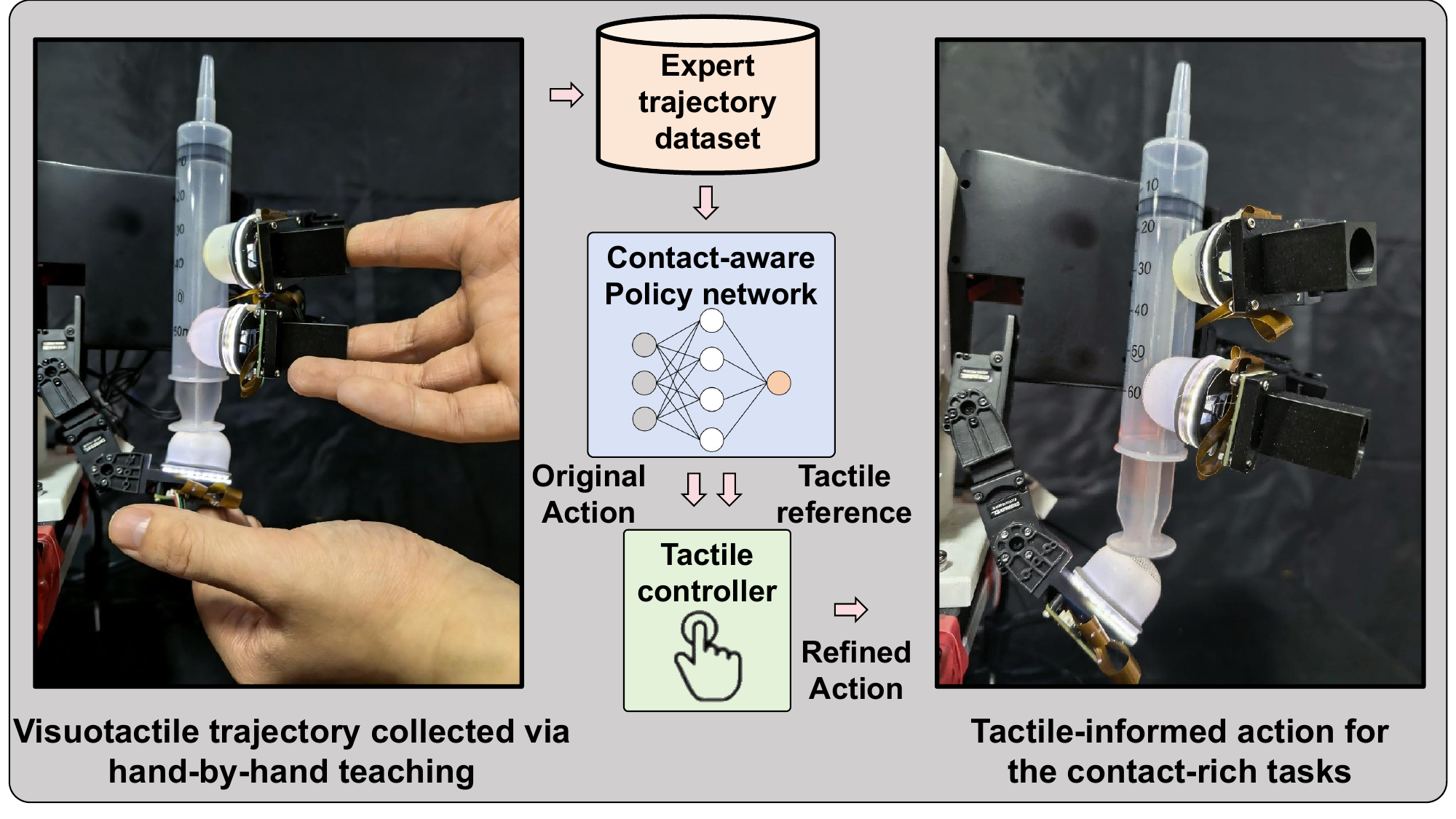}
	\caption{We present DexTac, a visuo-tactile manipulation learning framework via hand-by-hand teaching. It captures multi-dimensional tactile data to train the contact-aware policy network. With the tactile-informed motion generated by tactile controller, the dexterous hand can autonomously select the optimal contact region in the contact-rich tasks.}
	\label{fig:Syringe-pressing}
\end{figure}

Expert demonstrations in dexterous hand commonly employ teleoperation , which uses a VR headset \cite{Gallipoli} or data glove \cite{DexCap} to capture the expert's hand trajectories and maps them to the dexterous hand. However, this approach faces two critical limitations in contact-rich tasks: 1) kinematic mismatches between human and dexterous hands often prevent accurate reproduction of human motion trajectories \cite{Tilde}, and 2) the absence of tactile feedback during teleoperation makes task performance heavily depend on operator expertise \cite{Telerobotic}. Due to the above limitations, many recent studies have adopted kinesthetic teaching for expert demonstrations (a human directly manipulates the dexterous hand to perform a task)\cite{DexForce,Multimodal}. Prior research indicates that this method provides superior tactile interaction during demonstrations, significantly improving both teaching efficiency and success rates \cite{KineDex}. However, the above kinesthetic teaching methods on contact-rich tasks only collect low-dimensional tactile information-force, neglecting multi-dimensional tactile information about contact regions, which is important in the fine contact-rich manipulation task. 

To address the aforementioned challenges, we propose DexTac, which is shown in Fig. \ref{fig:Syringe-pressing}, a framework that leverages hand-by-hand teaching (kinesthetic teaching) to collect high-quality,  multi-dimensional visuotactile data and learns contact-aware policies from it. To meet the requirements of fine contact-rich manipulation tasks, our expert trajectory dataset incorporates dexterous hand joint information, visual data, and tactile feedback from terminal fingertips. Notably, in contrast to prior studies that predominantly capture low-dimensional tactile signals (e.g., contact forces), our approach systematically acquires high-dimensional tactile information, including contact force distributions and spatial contact regions.

However, even with a high-quality,  multi-dimensional visuotactile expert trajectory dataset, learning contact-aware policies from it via IL remains challenging. Most current IL research only focuses on predicting the robot's motion, while lacking prediction for tactile policies\cite{ACT,mobileACT,Diffusion,DP3}. However, in contact-rich tasks, policies that ignore tactile feedback result in low success rates. Only a few studies for contact-rich tasks incorporate predictions for tactile policies, yet these primarily concentrate on predicting force or motion trajectories modified based on force, neglecting the important factor of contact regions \cite{KineDex,DexForce,Multimodal}. In order to enable robot to apply right force within right contact regions, DexTac introduces both force and Center of Pressure (CoP), which represents the central contact region of the fingers, into the action prediction. Meanwhile, when deploying the policy network on the dexterous hand, the positions of the fingers are adjusted based on CoP error and predicted interaction forces to enable contact-aware manipulation.

Extensive experiments on dexterous syringe pressing demonstrate that our proposed method effectively guides the robotic fingers to engage the syringe using the correct contact regions. This approach achieves an 91.67\% average success rate across various syringe models, representing a 31.67\% relative improvement over the force-only baseline. Furthermore, the successful zero-shot transfer to a 20 mL syringe validates the strong generalization capability of our method.

The contributions of this paper are summarized as follows.

1) We propose DexTac, a framework for collecting multi-dimensional expert visuotactile trajectory via hand-by-hand teaching and learning multimodal contact-aware visuotactile policies from it via imitation learning (IL), which enables the dexterous hand to autonomously select and maintain optimal contact regions during contact-rich tasks.

2) The center of pressure (CoP) is introduced into the policy learning framework, which characterizes the centroid of the finger's contact region. In extensive experiments on dexterous syringe pressing tasks, incorporating CoP achieves a success rate of 91.67\%, representing a 31.67\% improvement over the force-based baseline method.

The rest of this paper is organized as follows. The related works are summarized in Section II. The proposed DexTac and contact-aware policies are introduced in Section III. In Section IV, the injection experiments are presented, also with the experimental results. Finally, this paper is summarized in Section V.
\section{RELATED WORKS} 
\subsection{Dexterous hand data collection}
Collecting high-quality expert trajectory dataset with multi-dimensional tactile data is essential for dexterous hand manipulation, especially in contact-rich tasks that demand precise physical interaction.
Most existing dexterous hand demonstrations collecting method employs teleoperation \cite{ACT,mobileACT,AnyTeleop,OPEN,Bunny-VisionPro}. However, as discussed in Section I, the human-robot kinematic mismatch and the absence of on-robot tactile feedback hinder both the efficiency and reliability of the data collection. To address this limitation, recent studies have proposed exoskeleton-based systems that deliver real-time haptic feedback as a surrogate for tactile sensation \cite{DOGlove,Immersive,SenseGlove}. Nevertheless, the resulting interaction experience remains markedly distinct from that of direct physical contact, and the efficiency of data collection is still on par with conventional teleoperation approaches. Empirical evidence from
prior work indicates that hand-by-hand teaching is preferred
over teleoperation, owing to its user-friendly nature and faster
demonstration acquisition \cite{DexForce,KineDex,demonstration,comparison,HIRO}. As demonstrated in Kinedex\cite{KineDex}, hand-by-hand teaching achieves a 4.7-fold reduction in demonstration time and elevates success rates from 46.7\% under teleoperation to 96.7\% for contact-rich tasks.

Motivated by the above dexterous hand data collection, DexTac employs hand-by-hand teaching to collect high-quality expert trajectory from directly from human demonstrations, which avoids the human-robot kinematic mismatch in the human hand motion retarget. Moreover, despite from the existing data collection methods only focus on the low-dimensional tactile information (e.g. contact force), our method captures multi-dimensional tactile data for the contact-aware policy learning.
\subsection{Contact-aware policies Learning}
With the high-quality visuotactile trajectory collected by DexTac, how to learn contact-aware policies from the data remains challenging. IL is a widely-used method to learn control policies from expert demonstrations, which enables embodied agents to learn autonomous manipulation policies in different tasks. However, the original IL method \cite{ACT,mobileACT,Diffusion,DP3} relies solely on robot joint information and visual data, lacking the integration of tactile information, which leads to lower success rates in contact-rich manipulation tasks. 

In robotic manipulation, existing works have incorporated tactile feedback into policy learning.
In \cite{Multimodal}, Ablett \emph{et al.} demonstrate the importace of tactile information through tactile force matching and tactile data as policy inputs in four door-opening and closing tasks. The results shows that force matching elevates the average policy success rate by 62.5\% and the integration of tactile data as a policy input leads to a 42.5\% enhancement. Besides, in the dexterous hand manipulation, DexForce \cite{DexForce} equips each fingertip with 6-axis force-torque sensors to record contact force vectors for the policy training and employ a impedance controller for multi contact-rich tasks. The most related work to ours is KineDex \cite{KineDex}, which employs soft silicone finger bands with embedded sensors to capture tactile information for policy training. However, similar to the above work, Kinedex only focuses on the contact force of the policy, without incorporating contact region into policy training. To satisfy the contact region requirements for fine, contact-rich manipulation tasks, we have further advanced the policy learning of KinDex by incorporating the learning of the CoP, which enables the dexterous hand to autonomously select contact regions. 
\section{METHODS}
\subsection{Overview of DexTac Framework}
\begin{figure*}
    \centering
    \includegraphics[width=0.99\linewidth]{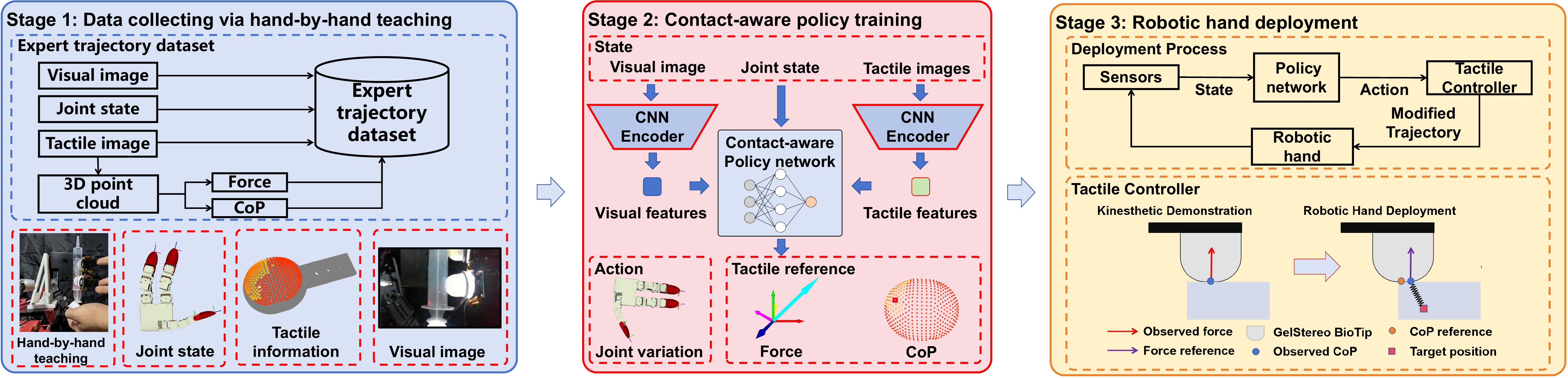}
    \caption{Overview of DexTac Framework. Stage 1: The expert trajectory dataset collected by hand-by-hand teaching, which  comprises visual images, tactile information, and joint state. Stage 2: the overview of the process of our contact-aware policy training. Stage 3: Top: the deployment process of our method. Bottom: the schematic diagram of position modification in the tactile controller}
    \label{fig:framework}
\end{figure*}

The overview of the DexTac framework is shown in Fig. \ref{fig:framework}. The objective of this framework is to learn contact-aware policies from visuotactile demonstrations based on hand-by-hand teaching for dexterous contact-rich tasks. Our pipeline consists of three main stages. First, we collect data via hand-by-hand teaching.  In this stage, to satisfy the demand of contact-rich tasks for high-quality tactile feedback, a dataset comprising dexterous hand joint states, RGB images, and multi-dimensional tactile information is collected through expert hand-by-hand teaching. Then, the contact-aware policy is trained based on the ACT framework, where joint angles, RGB images, and tactile images serve as state space, while joint variations, forces, and CoP constitute the action space. Last, in the deployment on the real-world dexterous hand, we employ the predicted forces and CoP to modify the robot motion, which can enable the hand to apply appropriate forces within correct contact regions for
successful manipulation. 
\begin{figure}
	\centering
	\includegraphics[width=0.99\linewidth]{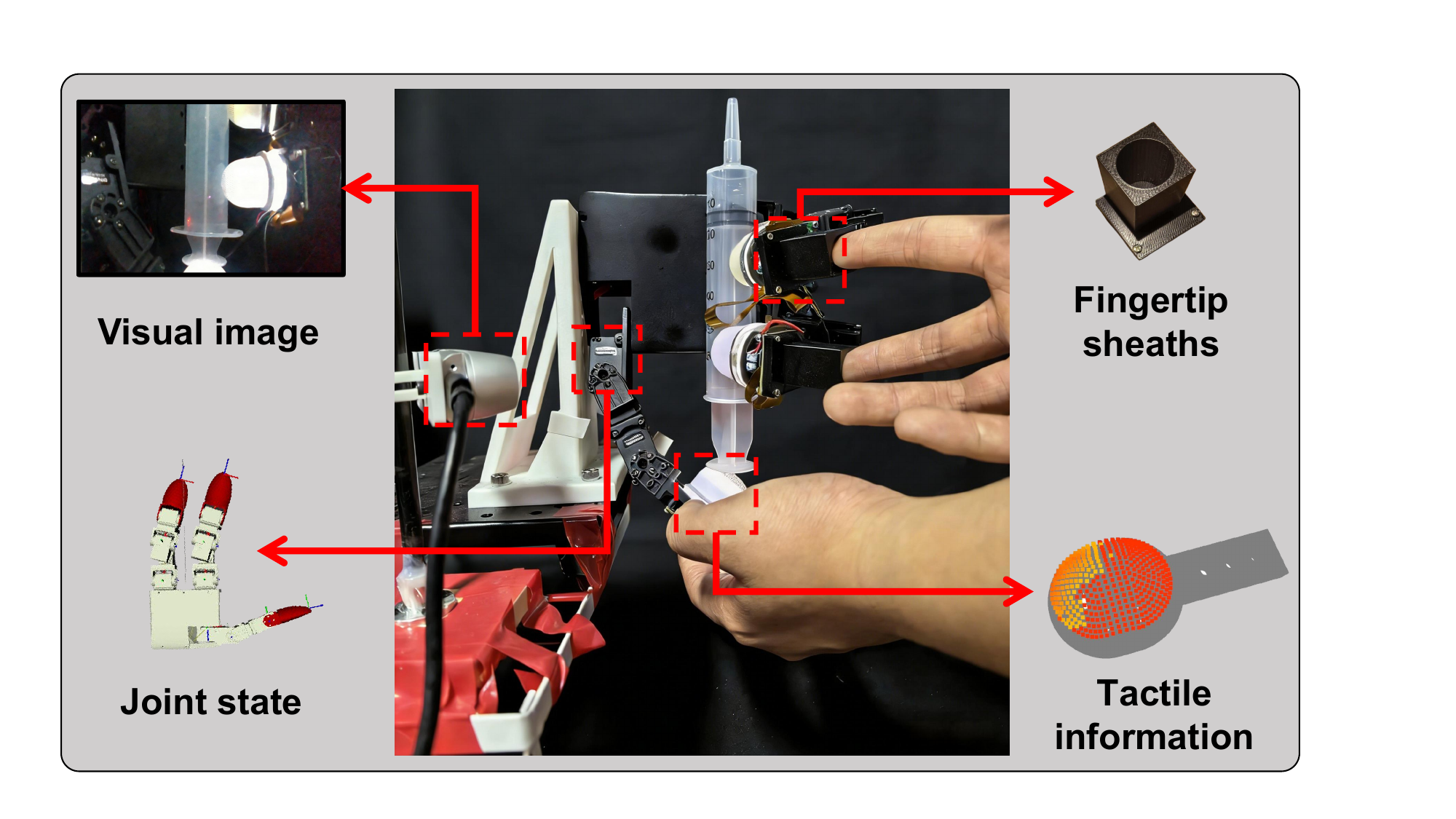}
	\caption{Hand-by-hand data collection. During data collection, the sheaths occlude the human fingers, which can address the problem of domain shift issues.}
	\label{fig:teaching}
\end{figure}
\subsection{Hand-by-hand Data Collection}
The core idea of hand-by-hand teaching is to allow a human directly manipulates the dexterous hand to perform the task. However, there are two challenges in this stage: (1) The kinematic mismatch between the human hand and the dexterous hand complicates the hand-by-hand teaching process. (2)  The presence of the human hand in visual images introduces domain shift issues in the collected data. To address the aforementioned issues, we design fingertip sheaths for our dexterous hand, as illustrated in Fig \ref{fig:teaching}. During data collection, these sheaths occlude the human fingers, and they remain on the robotic fingers during policy deployment. This design mitigates domain shift in the collected data and enables more intuitive and effective manipulation of the dexterous hand by the human operator, thereby improving both the efficiency and success rate of data acquisition.

We employ GelStereo BioTip sensors \cite{GelStereoBioTip} to capture tactile interactions. This sensor platform has been validated for interaction force \cite{IROS} and contact region estimation via visuotactile sensing. Each GelStereo BioTip sensor utilizes a stereo RGB camera pair to capture tactile images, which are then processed through 3D vision-based reconstruction to generate a tactile point cloud comprising $415$ points. The spatial configuration of this tactile point cloud is used to represent the acquired tactile information, which is shown in Fig. \ref{fig:teaching}.

Define $P_i(t)\in \mathbb{R}^3,i=1,2,...,415$ as the position of the $i$-th point.  
Then the displacement of the $i$-th point can be represented as 
\begin{equation}\label{eq1}
    \Delta P_i(t) = P_i(t)- P_i(0).
\end{equation}

To mitigate the influence of minor perturbations, in the GelStereo BioTip sensor, a deformation is considered a valid press only when the displacement exceeds a predefined threshold $\delta$. We define the following pressing magnitude accordingly:
\begin{equation}\label{eq2}
    \left\{ \begin{array}{l}
	y_i=1,\varDelta P_i>\delta,\\
	y_i=0,\varDelta P_i\le \delta.\\
\end{array} \right. 
\end{equation}
The interaction force is defined as the   displacement sum of the pressed points:
\begin{equation}\label{eq3}
    f_z = \sum_{i=1}^{415}{\Delta P_{i,z}y_i},
\end{equation}
where $\Delta P_{i,z}$ denotes the $z$-axis component of $\Delta P_{i}$, which means we only consider the force of the $z$-axis. And the fingertip CoP is defined as:
\begin{equation}\label{eq4}
\left\{ \begin{array}{l}
    c_t = \frac{1}{415}\sum_{i=1}^{415}{P_{i}}, y_i = 0\\
    c_t = \frac{1}{\sum_{i=1}^{415}y_i}\sum_{i=1}^{415}{P_{i}y_i}, y_i > 0.
    \end{array} \right. 
\end{equation}
In the stage of data collection, we only focus on $c_{t,xy}$, which are the $x$ and $y$-axis components of $c_t$.

During hand-by-hand teaching, the following data modalities are captured for each demonstration:
\begin{itemize}
    \item \textit{Visual observations}: a RGB image $o_t \in \mathbb{R}^{640\times  480\times 3}$ captured by the fixed camera.
    \item \textit{Tactile observations}: 3 tactile images $T_t  \in \mathbb{R}^{6\times 640\times  480\times 3}$ captured by 3 GelStereo BioTip sensors.
    \item \textit{Proprioception}: The robot hand’s joint states $q_t \in \mathbb{R}^{11}$ and its variations $\Delta q_t \in \mathbb{R}^{11}$.
    \item \textit{Interaction forces}: The interaction forces of 3 fingertips $F_z = [f_{z,1},f_{z,2},f_{z,3}]\in \mathbb{R}^{3}$ measured by 3 GelStereo BioTip sensors, where $f_{z,j}, j=1,2,3$ represents the $j$-th sensor's force.
    \item \textit{CoP}: The center of pressure of 3 fingertips $C_t = [c_{t,xy,1},c_{t,xy,2},c_{t,xy,3}] \in \mathbb{R}^{3\times 2}$ measured by 3 GelStereo BioTip sensors, where $c_{t,xy,j}, j=1,2,3$ represents the $j$-th sensor's CoP.
\end{itemize}
\begin{figure}
	\centering
	\includegraphics[width=0.9\linewidth]{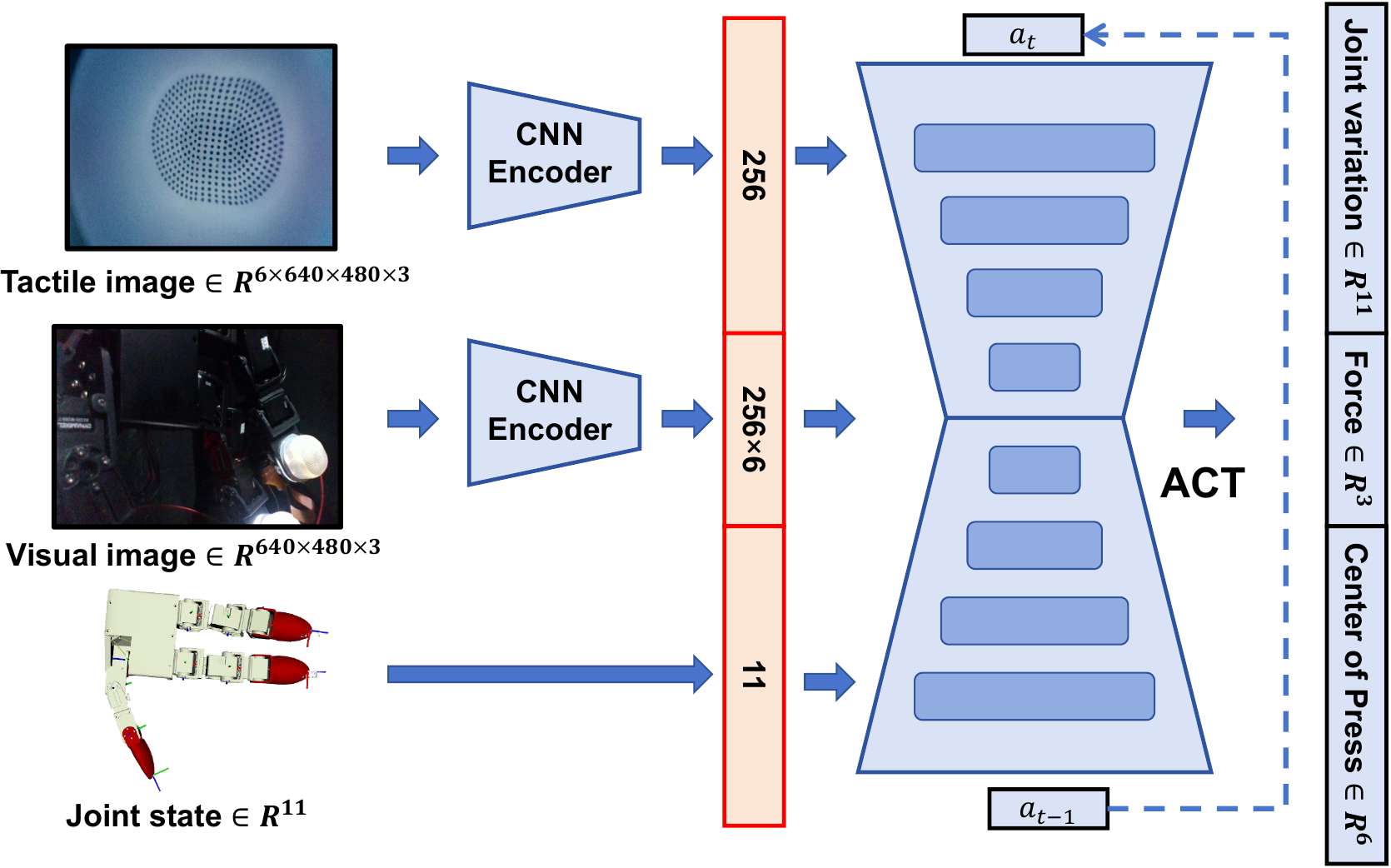}
	\caption{The structure of contact-aware policy network.The state space is composed of visual images, tactile images, and the joint states, while the action space consists of force, CoP, and joint variations.}
	\label{fig:network}
\end{figure}

\subsection{Policy Learning}\label{mypolicy}
In the contact-aware policy training, the dexterous contact-rich task is formulated as a Markov Decision Process (MDP). MDP can be represented by a tuple $\mathcal{M}=\left( \mathcal{S},\mathcal{A},\mathcal{P},\mathcal{R},\rho_{0} \right) $, where $\mathcal{S}$ and $\mathcal{A}$ are state space and action space, $\mathcal{P}$ is the state-transition environment dynamics distribution, $\mathcal{R}: \mathcal{S} \times \mathcal{A} \rightarrow \mathbb{R} $ is a reward function and $\rho_{0}$ is the initial state distribution. The policy $\pi$ interacts with the environment to generate a sequence of experiences $(s_t, a_t, r_t, s_{t+1})$ for $t = 0, \dots, T$, where the initial state is sampled as $s_0 \sim \rho_0(\cdot)$, the action is selected according to the policy $a_t = \pi(s_t)$, the next state follows the environment dynamics $s_{t+1} \sim P(\cdot \mid s_t, a_t)$, and the reward is given by $r_t = R(s_t, a_t)$. Here, $T$ denotes the finite horizon length. The trajectory starting from time step $t$ is denoted as $\tau_{t:T} = \{(s_t, a_t), \dots, (s_T, a_T)\}$. 

In this work, we focus on IL, a setting in which the reward function $R$ is unknown during training. Instead, the learner is provided with a finite dataset of task-specific expert demonstrations, denoted as $\mathcal{D}_E = \{(s, a), \dots\}$, consisting of state--action pairs generated by an expert policy. The objective of our work is to train a contact-aware policy $\pi$ to enable successful execution of the dexterous injection task.

To construct our policy network,  define the state space and action space as follows:
\begin{itemize}
    \item \textit{State space}: A RGB image $o_t$, three tactile images $T_t$ and the robot hand’s joint angles $q_t$.
    \item \textit{Action space}: The robot hand’s joint variations $\Delta q_t$ , the interaction forces of 3 fingertips $F_z$ and the CoP of 3 fingertips $C_t$. The action can be represented as $a_{t} = \{\Delta q_d,F_d,C_d\}$
\end{itemize}
The structure of our policy network is shown in Fig. \ref{fig:network}.

To train the contact-aware policy, we optimize the following loss:  
\begin{equation}\label{act}
\begin{aligned}
	\mathcal{L}_{\text{ACT}} &= \mathbb{E}\left[ \| a_{t:t+k} - \hat{a}_{t:t+k} \|_1 \right] \\
	&+ \beta \cdot D_{\mathrm{KL}}\left( q_{\phi}(z \mid s_t, a_{t:t+k}) \,\|\, \mathcal{N}(0, I) \right),
\end{aligned}
\end{equation}
where the first term minimizes the $L_1$ error between predicted and ground-truth action chunks $a_{t:t+k}$, ensuring precise trajectory matching, while the KL regularization term enforces the latent space $z$ to conform to a standard Gaussian prior, balancing reconstruction accuracy and latent disentanglement.

Our contact-aware policy produces action chunks during inference time for smoother control. Each action chunk contains desired joint variation, force and CoP, which will be applied in our tactile controller.
\begin{figure}
	\centering
	\includegraphics[width=0.9\linewidth]{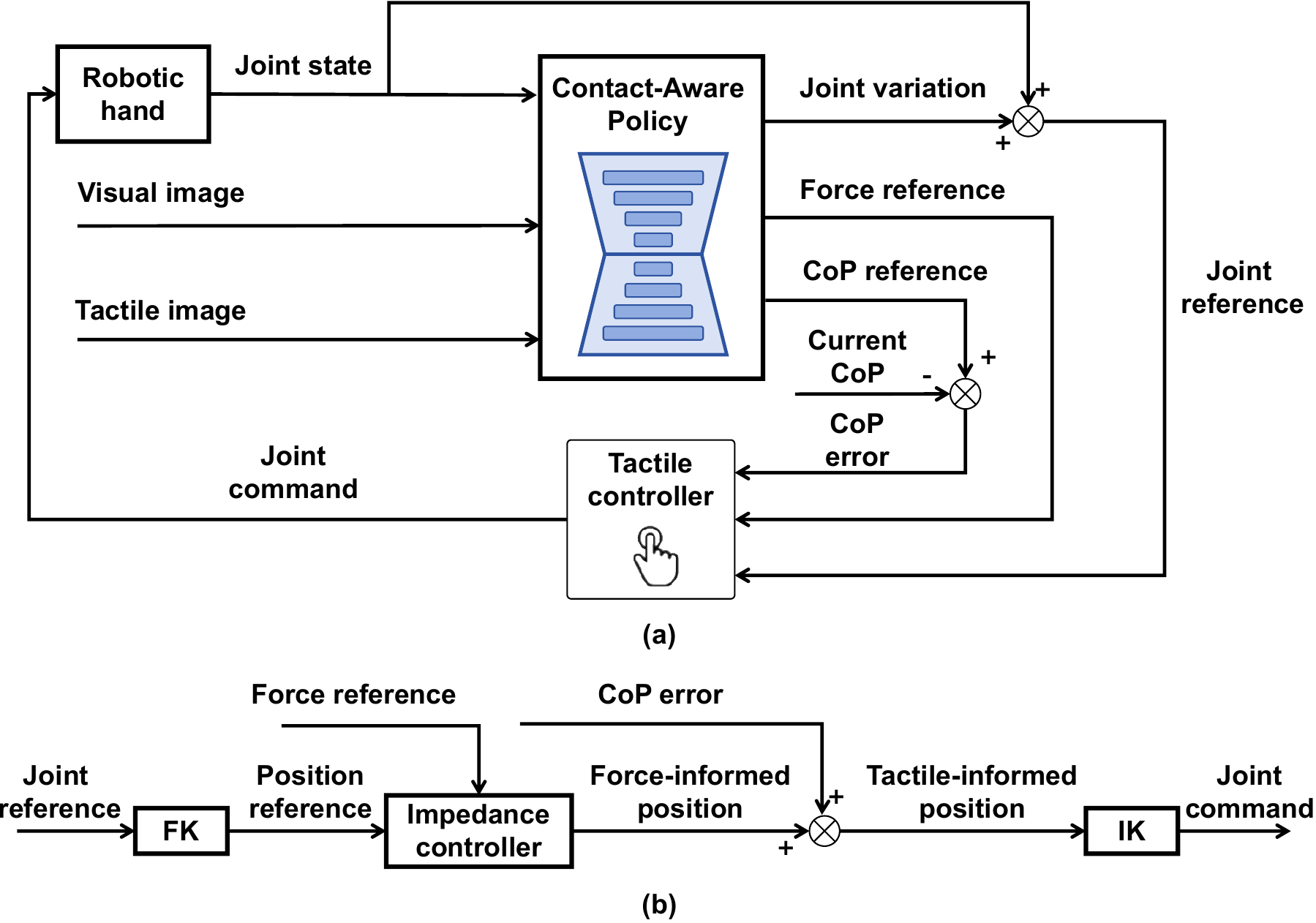}
	\caption{(a) The process of dexterous hand deployment. (b) The structure of tactile controller. The position reference is first corrected based on force, then further refined using CoP, before being output to the dexterous hand.}
	\label{fig:controller}
\end{figure}
\subsection{Dexterous Hand Deployment}
After the policy network learning, we deploy it on the dexterous hand system to perform contact-rich manipulation tasks. The deployment process is illustrated in Figure \ref{fig:controller} (a). Real-time sensory data including joint states, visual inputs, and tactile image are acquired through embedded sensors. The contact-aware policy network generated initial action and tactile references based on these multimodal inputs, which are then refined by the tactile controller to ultimately execute precise control of the dexterous hand.

In the traditional controller, robots are typically operated under position control \cite{siciliano2016springer}. However, in the contact-rich task, relying solely on position control may lead to task failure due to insufficient applied force or inaccurate contact region. To solve this problem, based on the tactile-informed action produced by the contact-aware policy, we propose a tactile controller for dexterous hand, which is shown in Fig. \ref{fig:controller} (b).

Recent studies usually introduce an impedance controller to apply appropriate force to the dexterous hand \cite{KineDex,DexForce,Multimodal}. Impedance control \cite{Impedance} formulates the interaction between force and motion by representing it as a second-order mechanical system, defined by specified parameters of mass, damping, and stiffness \cite{siciliano2016springer}. Rather than relying on explicit force control—which directly tracks desired forces—impedance control offers an alternative approach to simultaneously regulate both motion and interaction forces.

In the Cartesian impedance control, the control force $F\in \mathbb{R}^3$ for one finger is formulated as:
 \begin{equation}\label{force1}
 	\begin{aligned}
 		F = K(x_d-x)+D(\dot{x}_d-\dot{x})+M(\ddot{x}_d-\ddot{x}),
 	\end{aligned}
 \end{equation}
 where $x_d, \dot{x}_d, \ddot{x}_d \in \mathbb{R}^3$ denote the desired Cartesian position, velocity, and acceleration of the fingertip, respectively; $x, \dot{x}, \ddot{x} \in \mathbb{R}^3$ represent the actual values; and $K, D, M \in \mathbb{R}^{3 \times 3}$ are symmetric positive-definite matrices characterizing the desired stiffness, damping, and inertia properties of the impedance model. 
 
 To simplify the impedance model, we assume that $\dot{x}_d=\dot{x}$ and $\ddot{x}_d=\ddot{x}$. The Eq. (\ref{force1}) can be  simplified as follows:
 \begin{equation}\label{force}
 	\begin{aligned}
 		F = K(x_d-x).
 	\end{aligned}
 \end{equation}
  \begin{figure*}
 	\centering
 	\includegraphics[width=0.9\linewidth]{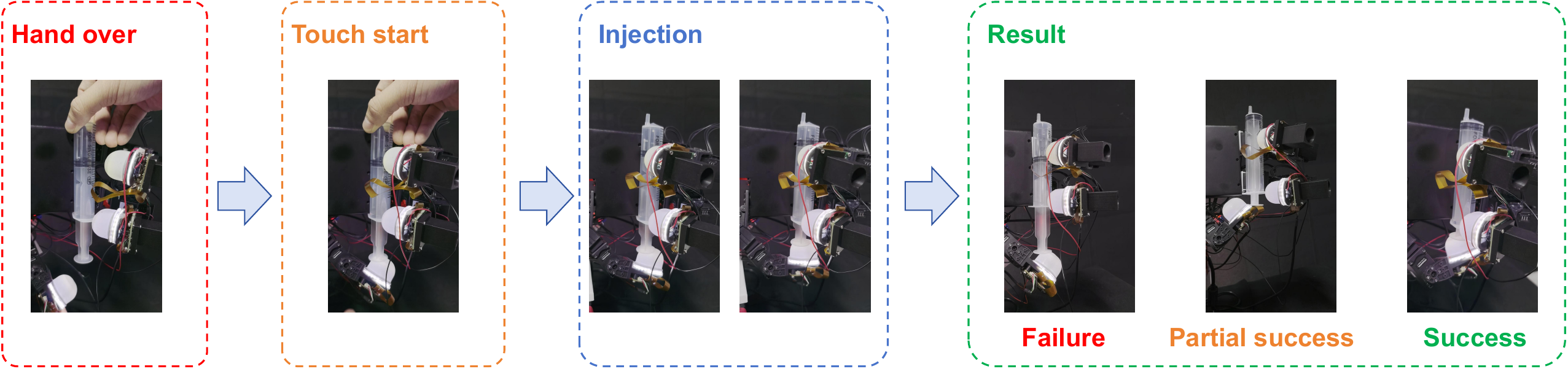}
 	\caption{The process and success criteria of injection task. The dexterous hand is required to first securely grasp a syringe from an initially ungrasped state and then fully depress the plunger using its thumb. The fingers are tasked with applying appropriate forces at suitable contact regions.}
 	\label{fig:setup}
 \end{figure*}  
 In our controller, the contact-aware policy network produces robot hand's joint variations $\Delta q_d$ and force of $z$-axis $F_z$. We can first obtain the position reference with the forward kinematics
  \begin{equation}\label{position}
 	\begin{aligned}
 		x_p = f(q_t+\Delta q_d),
 	\end{aligned}
 \end{equation}
 where $f(\cdot)$ denotes the forward kinematics function. $q_t$ denotes the actual joint angles. $x_p\in \mathbb{R}^9$ represents the Cartesian position reference of $3$ fingertips.
 
 Impedance control Eq. (\ref{force}) allows the robot hand to apply forces by tracking desired positions. Due to the contact force is predominantly attributed to the normal pressure acting along the finger's z-axis, the force-informed position  reference is formulated as:
 \begin{equation}\label{force-informed}
 	\begin{aligned}
 	\left\{ \begin{array}{l}
 		x_{f,xy}=x_{p,xy},\\
 		x_{f,z}=x_{p,z}+K^{-1}F_z,\\
 	\end{array} \right. 
 	\end{aligned}
 \end{equation}
 where $x_{f,xy}\in \mathbb{R}^6$ and $x_{f,z}\in \mathbb{R}^3$ denote the $xy$-axis and $z$-axis of the force-informed position reference, respectively. $x_{p,xy}$ and $x_{p,z}$ represent the $xy$-axis and $z$-axis of the position reference.
 
 With the above force-informed position reference, dexterous hand can apply right force to the object. However, how to enable dexterous hand to contact within right regions remains a challenge. Thereby, similar to the impedance control, we introduce CoP produced by our policy network to modify the position reference, which is given as:
 \begin{equation}\label{CoP-informed}
 	\begin{aligned}
 		\left\{ \begin{array}{l}
 			x_{r,xy}=x_{f,xy}+P(C_d-C_t),\\
 			x_{r,z}=x_{f,z},\\
 		\end{array} \right. 
 	\end{aligned}
 \end{equation}
 where $P\in \mathbb{R}^{6\times 6}$ denotes a hyperparameter to determine the stiffness of fingertips. Here, $x_r$ is the final tactile-informed position reference, which contains the force and CoP information. With this tactile controller, our dexterous hand autonomously selects optimal contact regions and appropriate force. The effect of our controller will be proved in the experiment.
 
 \section{EXPERIMENTS}

 \subsection{Experimental Setup and Design}
To evaluate the effectiveness of the proposed DexTac framework for dexterous contact-rich tasks, we conducts a series of injection experiments in the real-world dexterous hand.

 \textit{1) Experimental Setup:} In the injection task, the dexterous hand is required to first securely grasp a syringe from an initially ungrasped state and then fully depress the plunger using its thumb. In the stage of data collection, human experts effectively operate three robotic fingers using fingertip sheaths for demonstration. In the stage of dexterous hand deployment, the pose of the syringe is randomized across trials. The index and middle fingers are tasked with applying appropriate forces at suitable contact regions to firmly stabilize the syringe and prevent slippage. Simultaneously, the thumb had to exert a controlled force at an optimal location to effectively push the plunger while avoiding unintended slippage off the syringe barrel. This task not only requires the fingers to apply precise forces to the syringe but also demands the selection of appropriate contact regions to prevent slippage. To evaluate the effect of our methods, three evaluation outcomes are defined as follows: 
 \begin{itemize}
 	\item \textit{Failure}: The syringe is not actuated within 15 seconds or slipped from the grasp;
 	
 	\item \textit{Partial Success}: The syringe plunger is partially depressed but not fully actuated, or the thumb slipped off the syringe during the pushing motion; 
 	
    \item \textit{Success}: The syringe plunger is fully depressed to its limit without any finger losing contact with the syringe.
 \end{itemize}
 
 We refer the reader to Fig. \ref{fig:setup} for the process and success criteria of injection task.

\textbf{Hardware Setup}: Our platform is constructed on the GEX hand \cite{GEX}. Three GelStereo BioTip sensors \cite{GelStereoBioTip} are mounted on the fingertips of our dexterous hand to provide tactile information. A RealSense D435i camera is rigidly mounted adjacent to the dexterous hand to provide visual input to the policy network. Through this camera, the system could observe the syringe’s dimensions and pose, as well as the positions of the dexterous hand’s fingers.
\begin{figure*}
	\centering
	\includegraphics[width=0.9\linewidth]{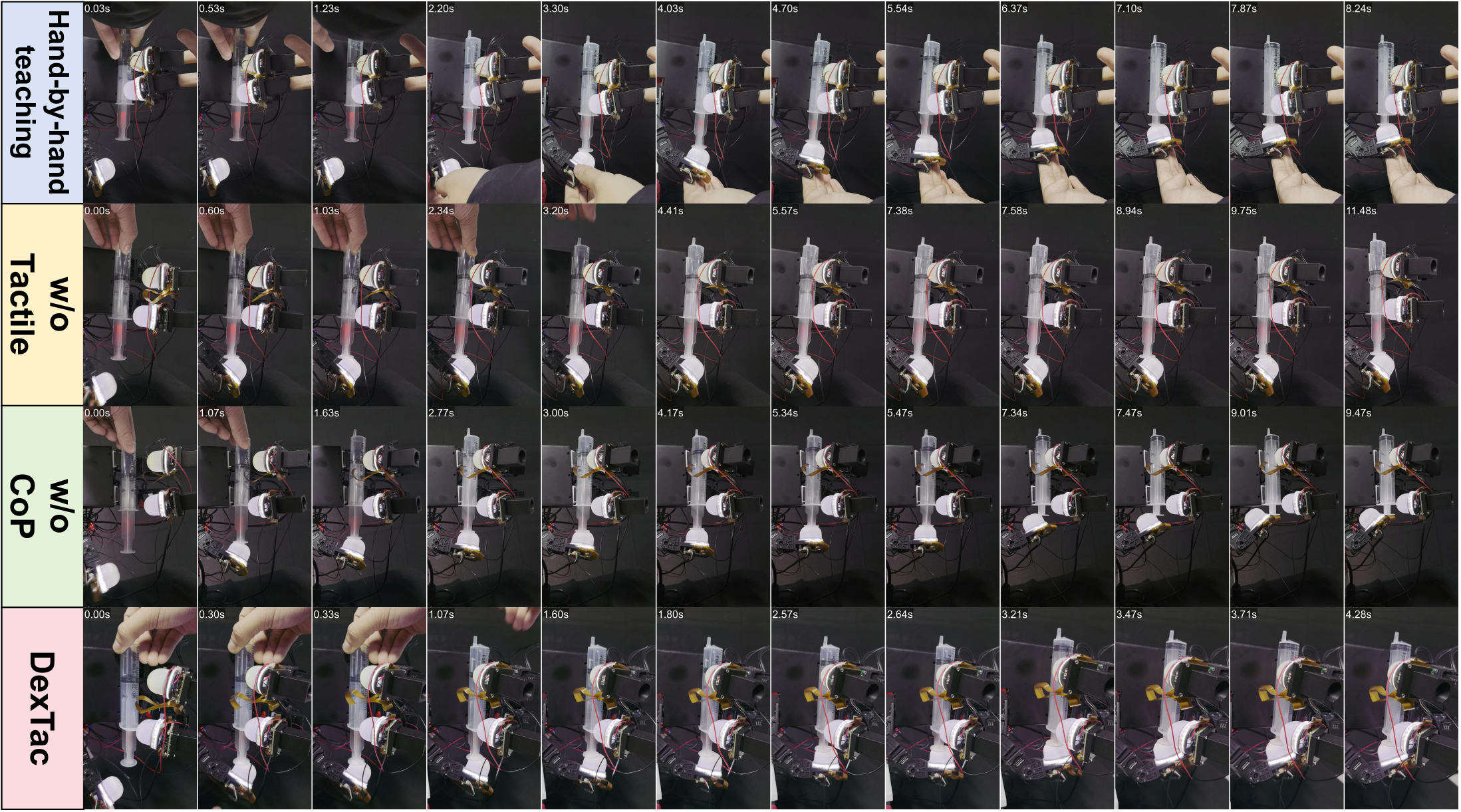}
	\caption{Example results for hand-by-hand teaching and three policies.}
	\label{fig:vedio}
\end{figure*}

\textbf{Parameters Training Setup}: Our policy network is built upon the ACT architecture, which processes multimodal sensory inputs through a shared CNN encoder based on ResNet-18 \cite{resnet} to extract features from one visual image and six tactile images (each at 640×480 resolution). The ResNet-18 backbone produces a 512-channel feature map, which is linearly projected to a 256-dimensional embedding space via a convolutional projection layer. These unified embeddings are then fed into the Transformer encoder to model long-horizon action sequences in a chunked manner, enabling coordinated visuo-tactile decision-making. Our policy is trained on Pytorch with the Adam optimizer with a learning rate of 0.0001. we record demonstrations at 30Hz and filter out data samples whenever the maximum angular change is less than 0.5 degrees. For each policy training process, we train the network for 2k gradient steps.

\textit{2) Experimental Design:} 
 To evaluate the effectiveness of the proposed DexTac framework for dexterous contact-rich tasks, we should answer the following questions:
\begin{itemize}
	\item Does the tactile reference, especially the CoP, improves the dexterous hand’s performance in the injection task?
	\item How well do our contact-aware policies generalize to out-of-distribution scenarios?
	\item How does the performance of our DexTac method depend on the amount of training data?
	\item Can DexTac be extended to purely tactile scenarios?
\end{itemize}

\textbf{Ablation study on DexTac:}
We conduct an ablation study on the tactile information mentioned in our method to investigate its impact on the overall performance. During experiments, we acquire 30 human demonstrations for each syringe size (30 mL, 40 mL, and 50 mL), totaling 90 demonstrations. All policies are trained on identical datasets but varied in terms of sensory modalities employed, with implementation details elaborated in subsequent subsections. Each policy is verified through 60 evaluation trials, with 20 trials allocated to each syringe size to ensure balanced performance assessment.

We design the following three comparative policies to demonstrate the importance of tactile sensing and the incorporation of CoP:
\begin{itemize}
	\item \textbf{w/o Tactile:} w/o Tactile: Following existing vision-based imitation learning paradigms \cite{ACT,Diffusion}, the model inputs consist solely of visual images and joint states, while disregarding the force and CoP corrections generated by the policy network during deployment on the controller.
	\item \textbf{w/o CoP:} Following KineDex \cite{KineDex} design paradigm, the model inputs are consistent with DexTac's configuration but omit CoP corrections during controller deployment.
	\item \textbf{DexTac:} Our proposed method.
\end{itemize}
\begin{figure*}
	\centering
	\includegraphics[width=0.99\linewidth]{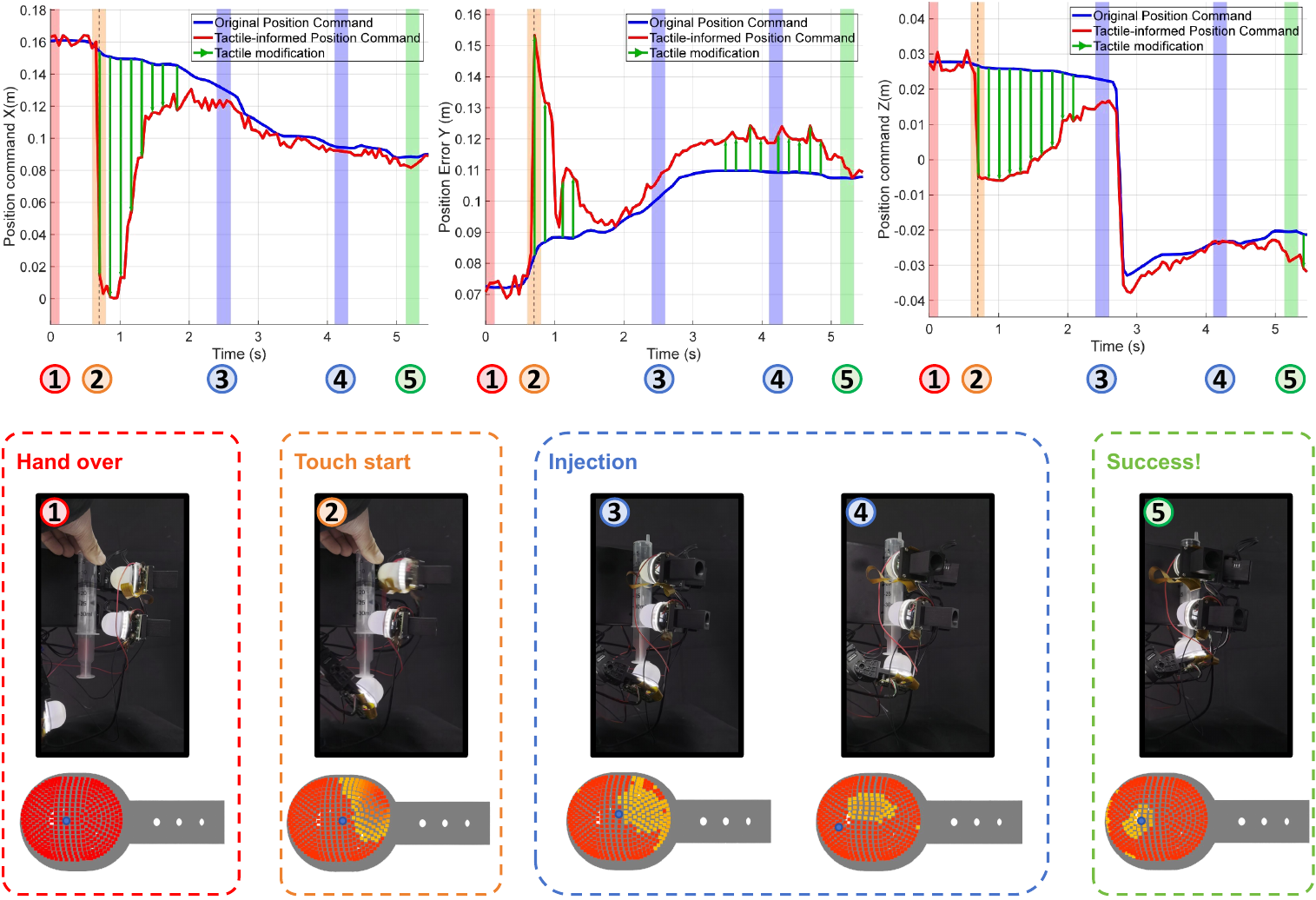}
	\caption{A successful example case result of DexTac. The first row of this figure shows the original position command (blue line), tactile modification (green arrow) and the tactile-modified position command. The second row shows the dexterous hand manipulation scene. The third row shows the tactile information and the CoP reference (blue point).The dexterous hand system continuously adjusts its contact region with the syringe based on the CoP error, ultimately accomplishing the task successfully while preventing finger slippage.}
	\label{fig:error}
\end{figure*}  
\begin{figure}
	\centering
	\includegraphics[width=0.99\linewidth]{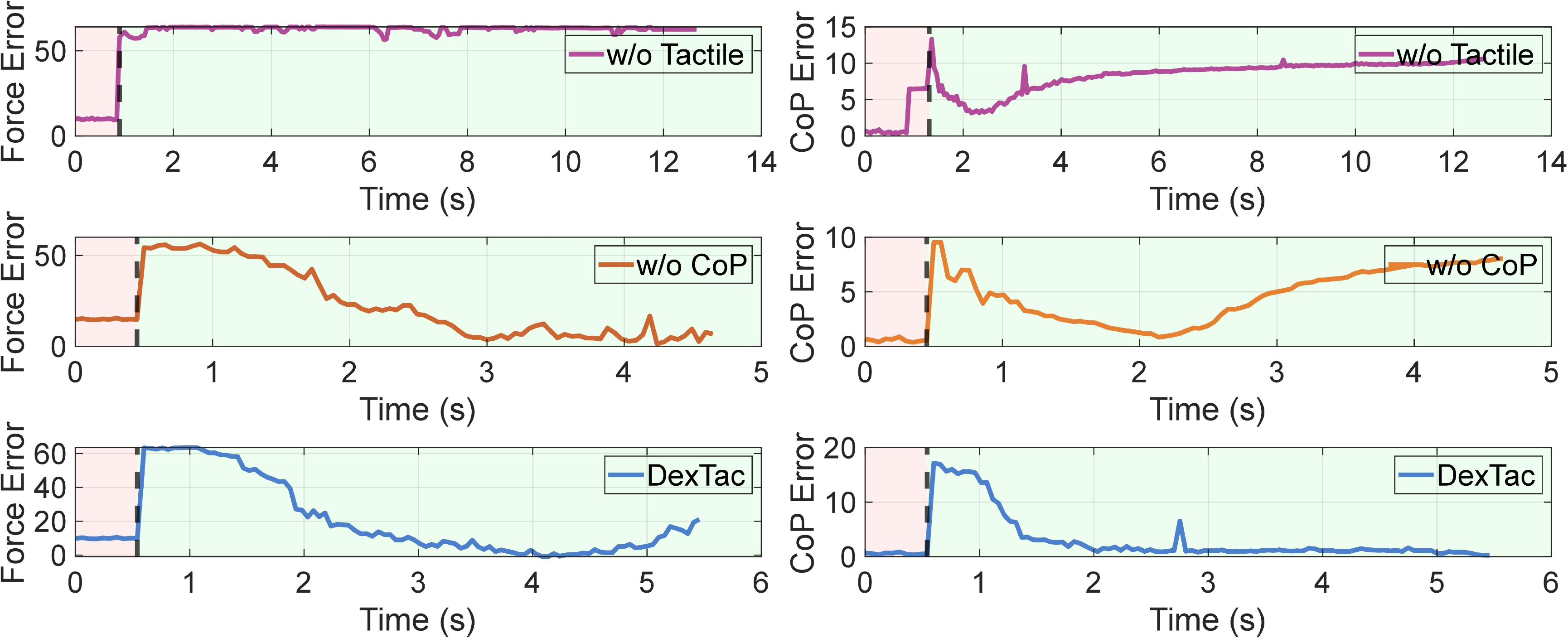}
	\caption{(a) and (b) The force error and CoP error of 3 policies where light red regions represent non-contact periods, and light green regions represent contact periods. Our DexTac method can ensure both the force error and CoP error remain convergent.}
	\label{fig:myerror}
\end{figure}

\textbf{Zero-shot transfer Experiment:} To evaluate the generalization capability of our methodology, we conduct direct deployment of the policy networks trained through ablation studies on a 20 mL injection task. Notably, the 20 mL syringe dimension is entirely unseen during the network training phase. The three policies identified in the ablation study are each evaluated through 20 independent trials to assess their experimental outcomes and compare their generalization capabilities.

\textbf{Data efficiency evaluation:}
To investigate the data efficiency of our methodology, we collect 50 human demonstration trajectories for each syringe size (20 mL, 30 mL, 50 mL, and 60 mL). The policy network is trained using progressively increasing subsets of the demonstration data: 10, 20, 30, 40, and 50 demonstrations per size. For each training subset, the corresponding task success rates are evaluated to quantify the relationship between data quantity and policy performance.

\textbf{Purely tactile experiment:} To verify the performance of our Dextac in the purely tactile scenarios. We conducted
two sets of experiments. In the experiment A, the fingers
initially do not grasp the syringe, requiring the system to first
establish a secure grip before proceeding to push the plunger.
In the experiment B, the fingers start already holding the
syringe, allowing the pushing action to be performed directly.
We performed 20 trials for each condition

\subsection{Experimental Results}
\textit{1) Ablation experiment:} 
 The example results of three policies are shown in the Fig. \ref{fig:vedio}. It can be observed that the method \textbf{w/o Tactile}
exhibits insufficient thumb pushing force or unstable grasping forces from the index and middle fingers. This often results in either failure to advance the syringe plunger or slippage of the syringe, significantly increasing the failure rate. Besides, the method \textbf{w/o CoP} is generally capable of securely grasping and actuating the syringe by applying appropriate interaction forces. However, the thumb tends to slip during plunger advancement since this policy lacks corrective feedback based on contact location. In our \textbf{DexTac} method, the policy network generate appropriate force and CoP for the injection task, which enables dexterous hand not only apply the right force to the syringe, but also autonomously select and maintain optimal contact regions. Therefore, the fingers remain firmly in place without slipping during the movement of the syringe, resulting in high success rate.
\begin{figure}
	\centering
	\includegraphics[width=0.99\linewidth]{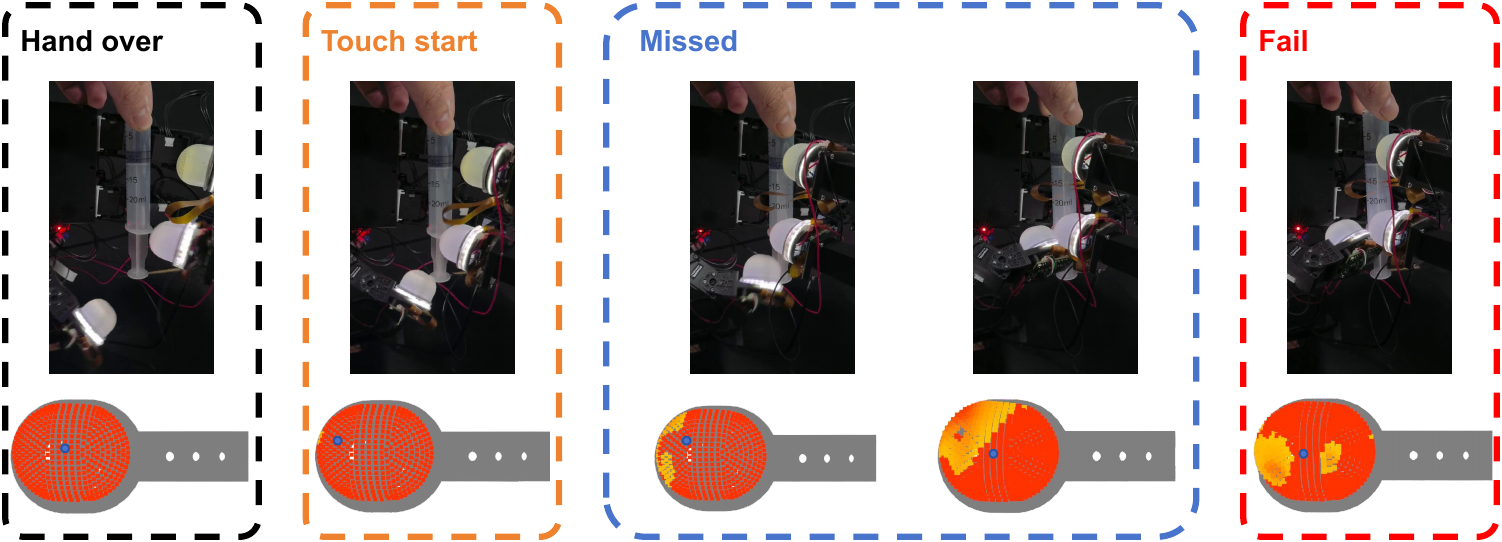}
	\caption{A failed example case result of DexTac. The first row shows the dexterous hand manipulation scene. The second row shows the tactile information and the CoP reference (blue point). Our method may grasp the syringe inaccurately when manipulating smaller syringes, which can lead to task failure.}
	\label{fig:fail}
\end{figure}  
\begin{table*}
	\centering
	\caption{The performance of each policy for injection task  across syringe sizes, where F, P and S represent Failure, Partial success and Success, respectively. DexTac demonstrated an overall success rate of 91.67\%, representing a statistically significant 31.67 \% improvement compared to the w/o CoP.}
	\label{tab:results}
	\begin{tabular}{l c c c c c c c c c c}
		\toprule
		\textbf{Policy} & \multicolumn{3}{c}{\textbf{30mL}} & \multicolumn{3}{c}{\textbf{50mL}} & \multicolumn{3}{c}{\textbf{60mL}} & \textbf{Success Rate (\%)} \\
		\cmidrule(lr){2-4} \cmidrule(lr){5-7} \cmidrule(lr){8-10}
		& \textbf{F} & \textbf{P} & \textbf{S} & \textbf{F} & \textbf{P} & \textbf{S} & \textbf{F} & \textbf{P} & \textbf{S} & \\
		\midrule
		w/o Tactile & 19 & 1 & 0 & 20 & 0 & 0 & 20 & 0 & 0 & 0.00 \\
		w/o CoP & 2 & 14 & 4 & 3 & 3 & 14 & 0 & 2 & 18 & 60.00 \\
		DexTac & 1 & 2 & 17 & 2 & 0 & 18 & 0 & 0 & 20 & 91.67 \\
		\bottomrule
	\end{tabular}
\end{table*}
\begin{figure*}
	\centering
	\includegraphics[width=0.99\linewidth]{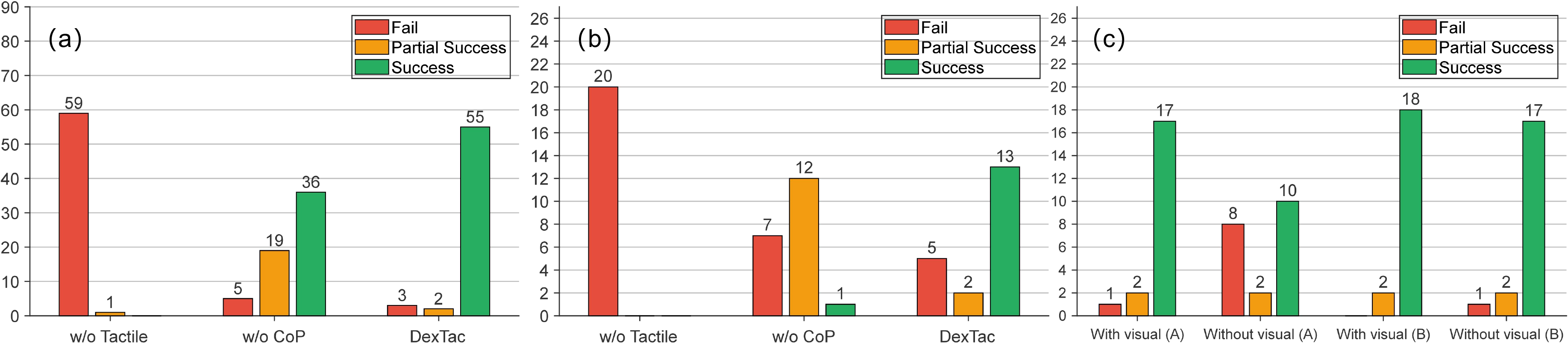}
	\caption{(a) The performance of different policies. DexTac shows a 31.67\% improvement in success rate compared with the force-only approach. (b) The performance of different syringe sizes. The 20 mL syringe is not trained in our policy network, but it still maintains a success rate of 65\% in the task. (c) The performance of input and experimental condition. In Experiment A, the absence of visual input lead to a 35\% decrease in the task's success rate. However, in Experiment B, the presence or absence of visual input had almost no effect on the task's success rate.}
	\label{fig:bar}
\end{figure*}  
\begin{table}
	\centering
	\caption{Zero-shot transfer experiment results across three control policies for 20 mL syringe.}
	\label{tab:zero}
	\begin{tabular}{l c c c}
		\toprule
		\textbf{Policy} & \textbf{Fail} & \textbf{Partial Success} & \textbf{Success} \\
		\midrule
		w/o Tactile       & 20            & 0                        & 0                \\
		w/o CoP           & 7             & 12                       & 1                \\
		DexTac            & 5             & 2                        & 13               \\
		\bottomrule
	\end{tabular}
\end{table}

A successful example case result of DexTac is shown in Fig. \ref{fig:error}, which presents a comparative example between the original position command and tactile-informed position command. By adjusting its initial position command, the system applies an appropriate force within the correct contact region of the syringe. From this example, we can observe that the dexterous hand system continuously adjusts its contact region with the syringe based on the CoP error, ultimately accomplishing the task successfully while preventing finger slippage. This example effectively illustrates how the dexterous hand accomplishes the task under our contact-aware policy.

To quantitatively demonstrate that our method enables the dexterous hand to apply appropriate forces within suitable contact regions, we recorded the force error and CoP error at the thumb’s fingertip. These errors are computed as the difference between the action outputs of the policy network and the real-time measurements from the fingertip sensor. The results are presented in Fig. \ref{fig:myerror}.
As shown in this figure, under the method \textbf{w/o tactile}, both the force error and the CoP error at the fingertip remain consistently high. This prevents the dexterous hand from applying the correct contact force at the appropriate location, ultimately leading to failure in the injection task. In contrast, under the method \textbf{w/o force}, the fingertip force error converges. However, the CoP error remains relatively large. Consequently, although the dexterous hand applies sufficient force during syringe manipulation, the lack of correction in contact location causes the thumb to gradually slip off the syringe, resulting in only partial success in the experiment. In contrast, under our proposed \textbf{DexTac} that incorporates CoP correction, both the force error and the CoP error converge to small values. This enables the dexterous hand to apply the appropriate contact force within a well-localized contact region, thereby ensuring successful completion of the injection task.

The performance of each policy for injection task is presented in the Table \ref{tab:results} and the Fig. \ref{fig:bar} (a).  It can be observed that the outcomes of method \textbf{w/o tactile} are predominantly clustered in the Failure category, while those of method \textbf{w/o CoP} are mostly concentrated in the Partial success category. In contrast, the results of \textbf{DexTac} are primarily concentrated in the Success category. \textbf{DexTac} demonstrated an overall success rate of 91.67\% across all experimental trials, representing a statistically significant 31.67 \% improvement compared to the \textbf{w/o CoP} baseline approach. In particular, compared to the original \textbf{w/o CoP} baseline approach, our method achieves an 85\% success rate in pushing a 30 mL syringe—a relatively small-sized syringe—representing a 65\% improvement in success rate. This demonstrates that our method exhibits significant advantages when manipulating smaller syringes.

\textit{2) Zero-shot transfer Experiment:} The zero-shot transfer performance results of $3$ policies are summarized in  Table \ref{tab:zero} and Fig. \ref{fig:bar} (b).
It can be observed that our method achieves a success rate of 65\% on the 20 mL syringe—even though no training data for this size is included—demonstrating its strong generalization capability. Besides, Compared to the \textbf{w/o CoP} approach, \textbf{DexTac} achieves a 60\% higher success rate in the 20 mL generalization experiments, further demonstrating the superiority and generalization capability of our method in fine manipulation tasks.

A fail example result is shown in Fig.\ref{fig:fail}. It can be observed that our method exhibits an issue of inaccurate initial grasping when manipulating smaller syringes, which can lead to task failure. This is likely attributable to the fact that the visual input in our approach is derived solely from a fixed camera, lacking multi-view visual information. Consequently, the imprecise localization of the syringe due to limited visual perspective may result in task failure.

\begin{figure}
	\centering
	\includegraphics[width=0.9\linewidth]{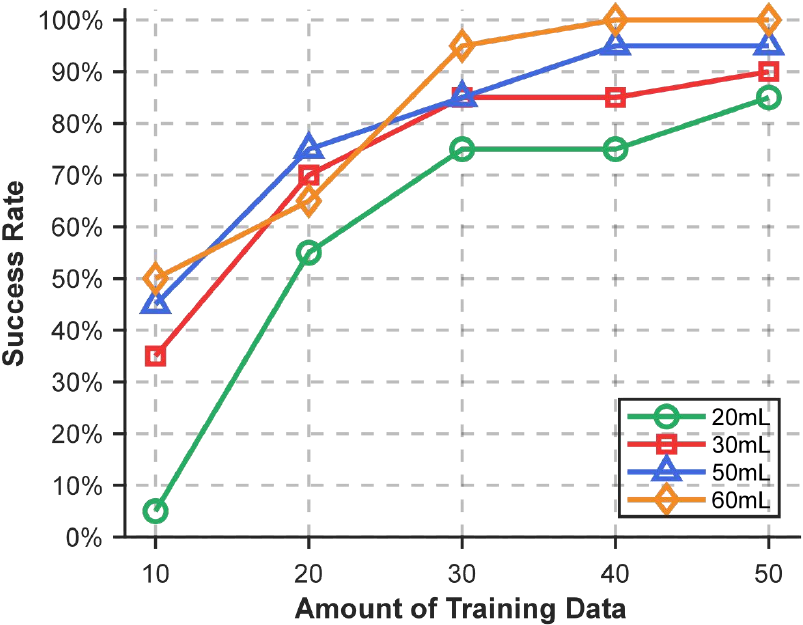}
	\caption{The performance of different data amounts. DexTac shows increasing performance with increasing data amounts, but the reach a plateau at 30 demonstrations for all syringe sizes.}
	\label{fig:amount}
\end{figure}

\textit{3) Data efficiency experiment:} The results are summarized in Fig. \ref{fig:amount}. It can be observed that, for all sizes of syringe, DexTac shows increasing performance with increasing data amounts, but the reach a plateau at 30 demonstrations.

\begin{figure}
	\centering
	\includegraphics[width=0.9\linewidth]{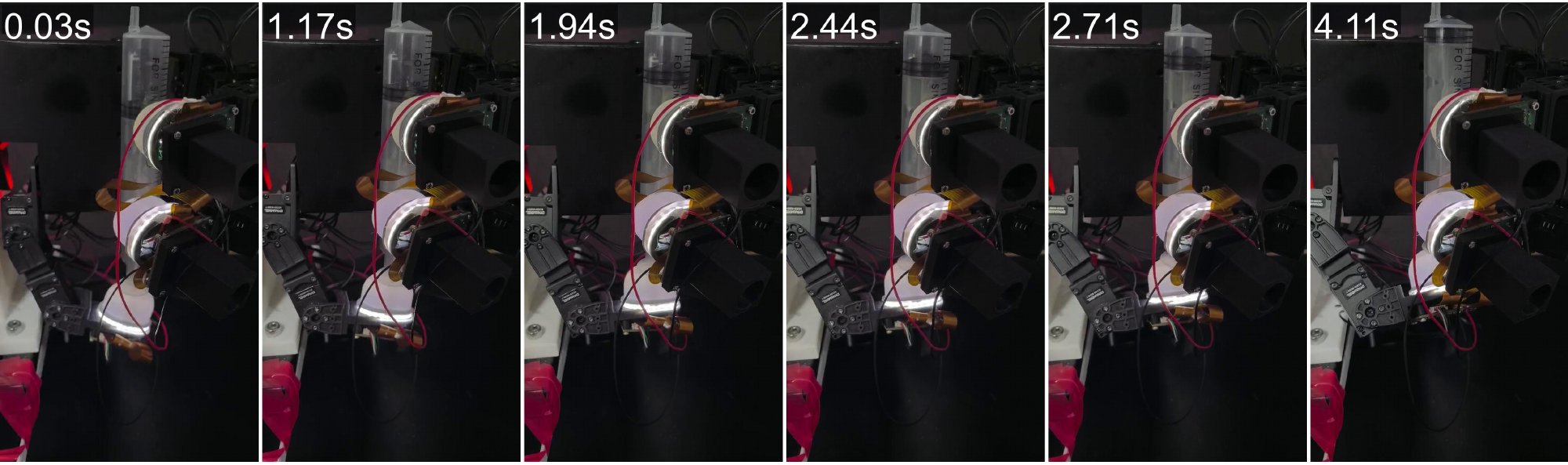}
	\caption{The example results of  purely tactile scenarios}
	\label{fig:initial}
\end{figure}

\textit{4) Purely tactile experiment:} The example results of  purely tactile scenarios are reported in Fig. \ref{fig:initial}. The performance of each condition is recorded in the Table \ref{tab:tactile} and Fig. \ref{fig:bar} (c), which shows that in Experiment
A, removing visual input significantly degrades task success.
In contrast, in Experiment B—where the fingers remain in continuous contact with the syringe throughout the task—the
absence of visual input has little impact on performance. This
suggests that, for tasks involving continuous physical contact,
our method can operate effectively without visual feedback,
offering a promising direction for such purely tactile scenarios.
\begin{table}
	\centering
	\caption{The performance of each  experimental condition}
	\label{tab:tactile}
	\begin{tabular}{l c c c}
		\toprule
		Condition & Fail & Partial Success & Success \\
		\midrule
		With visual (A) & 1 & 2 & 17 \\
		Without visual (A) & 8 & 2 & 10 \\
		With visual (B) & 0 & 2 & 18 \\
		Without visual (B) & 1 & 2 & 17 \\
		\bottomrule
	\end{tabular}
\end{table}

\section{CONCLUSION}
This paper proposes DexTac, a framework for collecting multi-dimensional expert visuotactile trajectory via hand-by-hand teaching and learning multimodal contact-aware visuotactile policies. To satisfy the contact region demands of fine contact-rich manipulation tasks,  DexTac captures expert trajectory dataset with multi-dimensional tactile data by integrating hand-by-hand teaching. Besides, we integrate rich modalities including force and CoP into the policy network built on ACT framework to learn contact-aware policies, allowing dexterous hand to autonomously select appropriate contact regions during manipulation. Then the contact-aware policy network is deployed into our dexterous hand system with a novel tactile controller. This controller regulates the dexterous hand's motion execution by incorporating tactile references—including contact forces and CoP—generated by the policy network, which enable our dexterous hand to apply appropriate forces within right contact regions.

Extensive experiments on dexterous syringe pressing validate the efficacy of DexTac. The framework achieves an average success rate of 91.67\% across diverse syringe models (e.g., 30 mL, 50 mL, and 60 mL), representing a 31.67\% relative improvement over force-only baselines. In zero-shot transfer to an unseen 20 mL syringe, DexTac attains a 65\% success rate, demonstrating strong generalization capability. These results underscore that incorporating CoP alongside force feedback significantly reduces contact region errors and slippage, particularly in high-precision scenarios with small-scale objects.

In the future work, we will explore extending DexTac to more complex deformable object manipulation, improving tactile calibration for sim-to-real transfer, and integrating additional sensory modalities for broader industrial applications. The framework’s effectiveness in contact-rich tasks paves the way for safer and more reliable dexterous manipulation.
\bibliographystyle{IEEEtran}
\bibliography{ref}
\end{document}